\documentclass[11pt,a4paper]{article}
\usepackage[hyperref]{naaclhlt2019}
\usepackage{times}
\usepackage{latexsym}
\usepackage{multirow}
\usepackage{paralist}
\usepackage{url}
\usepackage{algorithm}
\usepackage{algpseudocode}
\usepackage{graphicx}  
\usepackage{amsmath,CJK}

\aclfinalcopy 
\def\aclpaperid{1001} 

\title{Semantically-Aligned Equation Generation \\for Solving and Reasoning Math Word Problems}

 \author{Ting-Rui Chiang\quad Yun-Nung Chen \\
   National Taiwan University, 
   Taipei, Taiwan \\
   {\tt r07922052@csie.ntu.edu.tw \quad y.v.chen@ieee.org} \\}

\date{}

\begin{document}
\maketitle

\begin{abstract}
Solving math word problems is a challenging task that requires accurate natural language understanding to bridge natural language texts and math expressions.
Motivated by the intuition about how human generates the equations given the problem texts, this paper presents a neural approach to automatically solve math word problems by operating symbols according to their semantic meanings in texts.
This paper views the process of generating equations as a bridge between the semantic world and the symbolic world, where the proposed neural math solver is based on an encoder-decoder framework.
In the proposed model, the encoder is designed to understand the semantics of problems, and the decoder focuses on tracking semantic meanings of the generated symbols and then deciding which symbol to generate next.
The preliminary experiments are conducted in a benchmark dataset Math23K, and our model significantly outperforms both the state-of-the-art single model and the best non-retrieval-based model over about 10\% accuracy, demonstrating the effectiveness of bridging the symbolic and semantic worlds from math word problems.\footnote{The source code is available at \url{https://github.com/MiuLab/E2EMathSolver}.}
  
\end{abstract}

\section{Introduction}
Automatically solving math word problems has been an interesting research topic and also been viewed as a way of evaluating machines' ability~\cite{mandal2019solving}.
For human, writing down an equation that solves a math word problem requires the ability of reading comprehension, reasoning, and sometimes real world understanding.
Specifically, to solve a math word problem, we first need to know the goal of the given problem, then understand the semantic meaning of each numerical number in the problem, perform reasoning based on the comprehension in the previous step, and finally decide what to write in the equation.

Most prior work about solving math word problems relied on hand-crafted features, which required more human knowledge.
Because those features are often in the lexical level, it is not clear whether machines really understand the math problems.
Also, most prior work evaluated their approaches on relatively small datasets, and the capability of generalization is concerned.

\begin{figure*}
\centering
  \includegraphics[width=\linewidth]{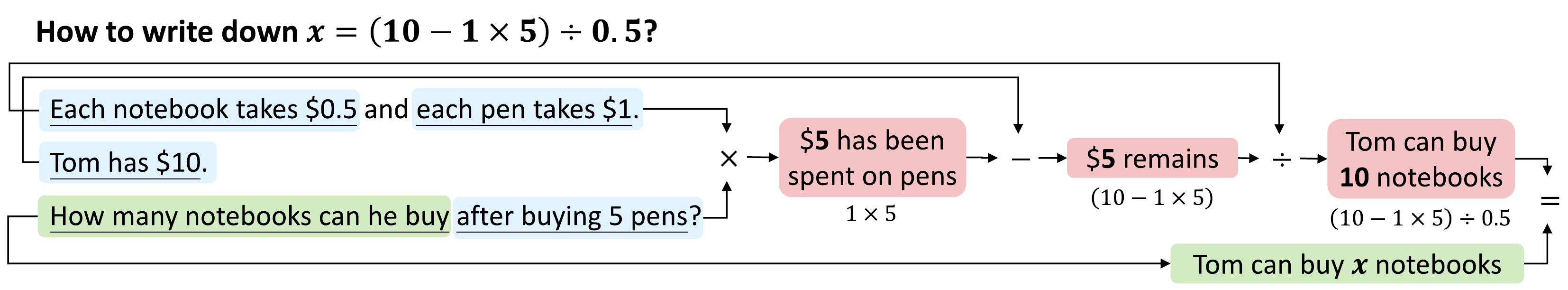}
\caption{The solving process of the math word problem ``\textit{Each notebok takes \$0.5 and each pen takes \$1. Tom has \$10. How many notebook can he buy after buying 5 pens?}'' and the associated equation is $x=(10-1\times 5)\div 0.5$. The associated equation is $x=(10-1\times 5)\div 0.5$.}
  \label{fig:solving}
\end{figure*}

This paper considers the reasoning procedure when writing down the associated equation given a problem.
Figure~\ref{fig:solving} illustrates the problem solving process.
The illustration shows that human actually assigns the semantic meaning to each number when manipulating symbols, including operands (numbers) and operators ($+ - \times \div$).
Also, we believe that the semantic meaning of operands can help us decide which operator to use. 
For example, the summation of ``\emph{price of one pen}'' and ``\emph{number of pens Tom bought}'' is meaningless; therefore the addition would not be chosen.

Following the observation above, this paper proposes a novel encoder decoder model, where the encoder extracts semantic meanings of numbers in the problem, and the decoder is equipped with a stack that facilitates tracking the semantic meanings of operands. 
The contributions of this paper are 4-fold:
\begin{itemize}
\item This paper is the first work that models semantic meanings of operands and operators for math word problems.
\item This paper proposes an end-to-end neural math solver with a novel decoding process that utilizes the stack to generate associated equations.
\item This paper achieves the state-of-the-art performance on the large benchmark dataset Math23K.
\item This paper is capable of providing interpretation and reasoning for the math word problem solving procedure.
\end{itemize}

\section{Related Work}

There is a lot of prior work that utilized hand-crafted features, such as POS tags, paths in the dependency trees, keywords, etc., to allow the model to focus on the quantities in the problems~\cite{kushman2014learning,hosseini2014learning,roy2015reasoning,roy2015solving,koncel2015parsing,roy2016equation,upadhyay2016learning,upadhyay2017annotating,roy2018mapping,wang2018mathdqn}.
Recently, \citeauthor{mehta2017deep,wang2017deep,ling2017program} attempted at learning models without predefined features.
Following the recent trend, the proposed end-to-end model in this paper does not need any hand-crafted features.

\citeauthor{kushman2014learning} first extracted templates about math expressions from the training answers, and then trained models to select templates and map quantities in the problem to the slots in the template.
Such two-stage approach has been tried and achieved good results~\cite{upadhyay2017annotating}.
The prior work highly relied on human knowledge, where they parsed problems into equations by choosing the expression tree with the highest score calculated by an operator classifier, working on a hand-crafted ``trigger list'' containing quantities and noun phrases in the problem, or utilizing features extracted from text spans~\cite{roy2015reasoning,roy2016equation,koncel2015parsing}.
\citeauthor{shi2015automatically} defined a Dolphin language to connect math word problems and logical forms, and generated rules to parse math word problems. 
\citeauthor{upadhyay2016learning} parsed math word problems without explicit equation annotations.
\citeauthor{roy2018mapping} classified math word problems into 4 types and used rules to decide the operators accordingly. \citeauthor{wang2018mathdqn} trained the parser using reinforcement learning with hand-crafted features.
\citeauthor{hosseini2014learning} modeled the problem text as transition of world states, and the equation is generated as the world states changing.
Our work uses a similar intuition, but hand-crafted features are not required and our model can be trained in an end-to-end manner.
Some end-to-end approaches have been proposed, such as generating equations directly via a seq2seq model~\cite{wang2017deep}. 
\citeauthor{ling2017program} tried to generate solutions along with its rationals with a seq2seq-like model for better interpretability.

This paper belongs to the end-to-end category, but different from the previous work; we are the first approach that generates equations with stack actions, which facilitate us to simulate the way how human solves problems.
Furthermore, the proposed approach is the first model that is more interpretable and provides reasoning steps without the need of rational annotations.

\begin{figure*}[t!]
  \centering
  \includegraphics[width=\linewidth]{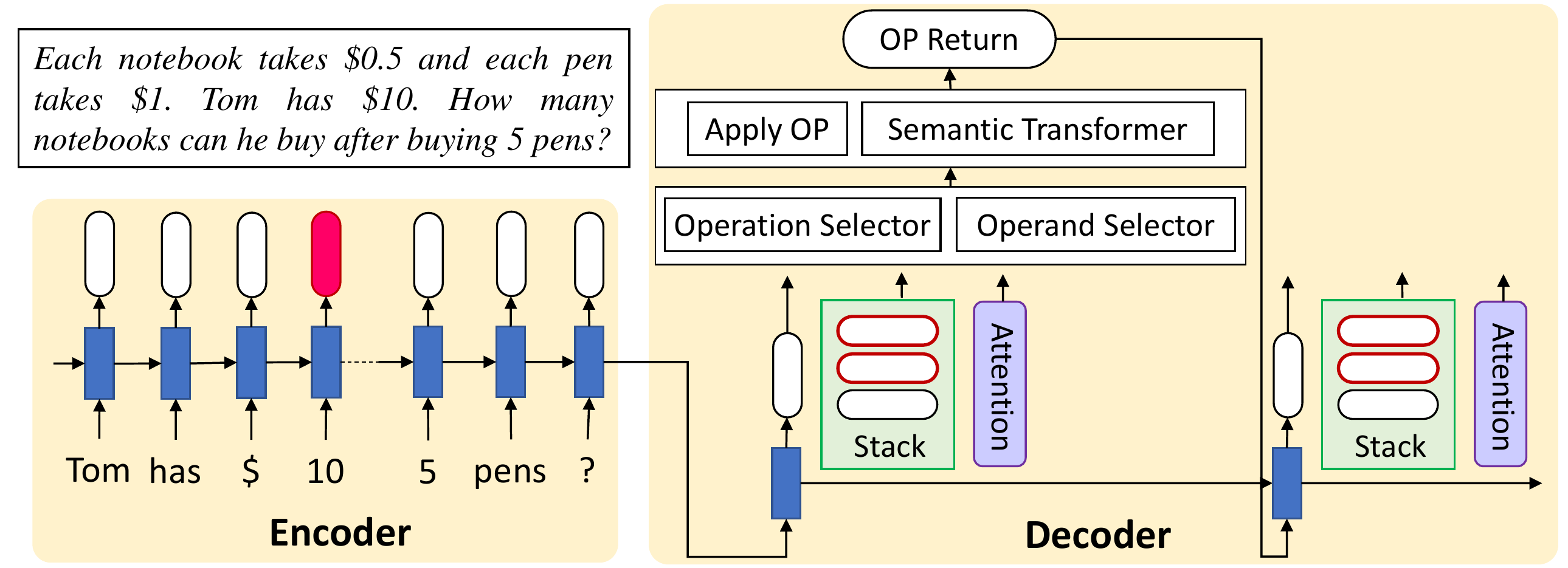}
  \caption{The encoder-decoder model architecture of the proposed neural solver machine.}
  \label{fig:model}
\end{figure*}

\section{End-to-End Neural Math Solver}

Our approach composes of two parts, an encoder and a decoder, where the process of solving math word problems is viewed as transforming multiple text spans from the problems into the target information the problems ask for.
In the example shown in Figure~\ref{fig:solving}, all numbers in the problem are attached with the associated semantics.
Motivated by the observation, we design an encoder to extract the semantic representation of each number in the problem text. 
Considering that human usually manipulates those numbers and operators (such as addition, subtraction, etc.) based on their semantics for problem solving, a decoder is designed to construct the equation, where the semantics is aligned with the representations extracted by the encoder.
The idea of the proposed model is to imitate the human reasoning process for solving math word problems.
The model architecture is illustrated in Figure~\ref{fig:model}.

\subsection{Encoder}

The encoder aims to extract the semantic representation of each constant needed for solving problems. 
However, the needed constants may come from either the given problem texts or domain knowledge, so we detail these two procedures as follows.

\subsubsection{Constant Representation Extraction}
For each math word problem, we are given a passage consisting of words $\{w_t^P\}_{t=1}^m$, whose word embeddings are $\{e_t^P\}_{t=1}^m$. 
The problem text includes some numbers, which we refer as constants. The positions of constants in the problem text are denoted as $\{p_i\}_{i=1}^n$. 
In order to capture the semantic representation of each constant by considering its contexts, a bidirectional long short-term memory (BLSTM) is adopted as the encoder~\cite{hochreiter1997long}:
\begin{equation}
  h_t^E, c_t^E = \mathrm{BLSTM}(h_{t-1}^E, c_{t-1}^E, e_{t}^P),
\end{equation}
and then for the $i$-th constant in the problem, its semantic representation $e^c_i$ is modeled by the corresponding BLSTM output vector:
\begin{equation}
  e^c_i = h^E_{p_i}.
  \label{eq:semantic-representation}
\end{equation}

\subsubsection{External Constant Leveraging} 
External constants, including $1$ and $\pi$, are leveraged, because they are required to solve a math word problem, but not mentioned in the problem text. Due to their absence from the problem text, we cannot extract their semantic meanings by BLSTM in (\ref{eq:semantic-representation}).
Instead, we model their semantic representation $e^{\pi}, e^{1}$ as parts of the model parameters.
They are randomly initialized and are learned during model training.


\subsection{Decoder}
The decoder aims at constructing the equation that can solve the given problem. 
We generate the equation by applying stack actions on a stack to mimic the way how human understands an equation.
Human knows the semantic meaning of each term in the equation, even compositing of operands and operators like the term "$(10 - 1 \times 5)$" in Figure \ref{fig:solving}.
Then what operator to apply on a pair operands can be chosen based on their semantic meanings accordingly. 
Hence we design our model to generate the equation in a postfix manner: a operator is chosen base on the semantic representations of two operands the operator is going to apply to. 
Note that the operands a operator can apply to can be any results generated previously. 
That is the reason why we use ``stack'' as our data structure in order to keep track of the operands a operator is going to apply to. 
The stack contains both symbolic and semantic representations of operands, denoted as
\begin{equation}
  S = [(v^S_{l_t}, e^S_{l_t}), (v^S_{l_t-1}, e^S_{l_t-1}), \cdots, (v^S_1, e^S_1)],
  \label{eq:stack}
\end{equation}
where $v^S$ of each pair is the symbolic part, such as $x + 1$, while $e^S$ is the semantic representation, which is a vector.
The components in the decoder are shown in the right part of Figure~\ref{fig:model}, each of which is detailed below.

\begin{figure*}[t!]
  \centering
  \includegraphics[width=0.96\linewidth]{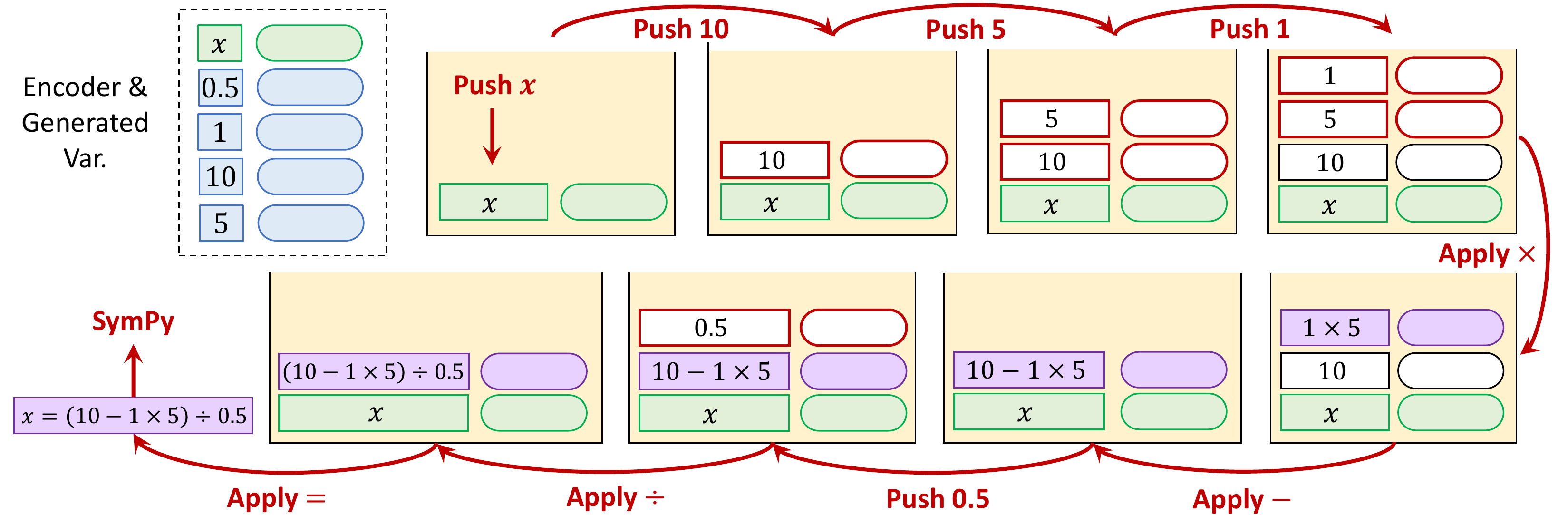}
  \caption{Illustration of the inference process. The purple round blocks denote the transformed semantics, while the green ones are generated by the variable generator.}
  \label{fig:infer}
\end{figure*}

\subsection{Decoding State Features}
\label{sec:decoding-state-features}
At each decoding step, decisions are made based on features of the current state. At each step, features $r^{sa}$ and $r^{opd}$ are extracted to select a stack action (section \ref{sec:stack-actions}) and an operand to push (section \ref{sec:operand-selector}). Specifically, the features are the gated concatenation of following vectors:
\begin{itemize}
    \item $h^D_{t}$ is the output of an LSTM, which encodes the history of applied actions:
    \begin{equation}
        h^D_t, c^D_t = \mathrm{LSTM}(h^D_{t-1}, c^D_{t-1},     \mathrm{res}_{t-1}),
        \label{eq:LSTM}
    \end{equation}
    where $\mathrm{res}_{t-1}$ is the result from the previous stack action similar to the seq2seq model \cite{sutskever2014sequence}.
    For example, if the previous stack action $o_{t-1}$ is ``push'', then $\mathrm{res}_{t-1}$ is the semantic representation pushed into the stack.
    If the previous stack action $o_{t-1}$ is to apply an operator $\diamond$, then $\mathrm{res}_{t-1}$ is the semantic representation generated by $f_{\diamond}$.

    \item $s_{t}$ is the stack status. It is crucial because some operators are only applicable to certain combinations of operand semantics, which is similar to the type system in programming languages.
    For example, operating multiplication is applicable to the combination of ``\textit{quantity of an item}'' and ``\textit{price of an item}'', while operating addition is not.
    Considering that all math operators supported here  ($+,-,\times,\div$) are binary operators, the semantic representations of the stack's top 2 elements at the time $t - 1$ are considered:
\begin{equation}
    s_t = [e^S_{l_t}; e^S_{l_t}].
\end{equation}

\item $q_{t}$ incorporates problem information in the decision. 
It is believed that the attention mechanism \cite{luong2015effective} can effectively capture dependency for longer distance.
Thus, the attention mechanism over the encoding problem $h^E_1, h^E_2, \cdots$ is adopted:
    \begin{equation}
        q_t = \mathrm{Attention}(h^D_t, \{h_i^E\}_{i=1}^m),
        \label{eq:decoding-attention}
    \end{equation}
    where the attention function in this paper is defined as a function with learnable parameters $w, W, b$:
    \begin{align}
        &\mathrm{Attention}(u, \{v_i\}_{i=1}^m) = \sum_{i=1}^m \alpha_i h_i, \\
        &\alpha_i = \frac{\exp(s_i)}{\sum_{l=1}^m \exp(s_i)}, 
        \label{eq:attention-weight}
        \\
        &s_i = w^\mathsf{T} \tanh (W^T[u; v_i] + b).
    \end{align}
\end{itemize}

In order to model the dynamic features for different decoding steps, features in $r_t^{sa}$ is gated as follows:
\begin{align}
  & r_t^{sa} = [g_{t,1}^{sa} \cdot h^D_t; g_{t,2}^{sa} \cdot s_t; g^{sa}_{t,3} \cdot q_t], \\
  & g_t^{sa} = \sigma(W^{sa} \cdot [h^D_t; s_t; q_t]),
  \label{eq: operator-gate}
\end{align}
where $\sigma$ is a sigmoid function and $W^{sa}$ is a learned gating parameter. $r_t^{opd}$ is defined similarly, but with a different learned gating parameter $W^{opd}$.

\subsubsection{Stack Action Selector}
The stack action selector is to select an stack action at each decoding step (section \ref{sec:stack-actions}) until the unknowns are solved. The probability of choosing action $a$ at the decoding step $t$ is calculated with a network $\mathrm{NN}$ constituted of one hidden layer and ReLU as the activation function:
\begin{align}
    P(Y_t \mid & \{ y_i \}_{i=1}^{t-1}, \{ w_i \}_{i=1}^m) \\ 
  = &\ \mathrm{StackActionSelector}(r^{sa}_t) \nonumber\\
  = &\ \mathrm{softmax}(\mathrm{NN}(r^{sa}_t)),\nonumber
\end{align}
where $r^{sa}_t$ is decoding state features as defined in section \ref{sec:decoding-state-features}.

\subsubsection{Stack Actions}
\label{sec:stack-actions}
The available stack actions are listed below:

\begin{itemize}
\item \textbf{Variable generation}: The semantic representation of an unknown variable $x$ is generated dynamically as the first action in the decoding process. 
Note that this procedure provides the flexibility of solving problems with more than one unknown variables.
The decoder module can decide how many unknown variables are required to solve the problem, and the semantic representation of the unknown variable is generated with an attention mechanism:
\begin{align}
  e^x = \mathrm{Attention}(h^D_t, \{h_i^E\}_{i=1}^m).
\end{align}

\item \textbf{Push}: 
This stack action pushes the operand chosen by the operand selector (section \ref{sec:operand-selector}).
Both the symbolic representation $v_*$ and semantic representation $e_*$ of the chosen operand would be pushed to the stack $S$ in (\ref{eq:stack}).
Then the stack state becomes
\begin{equation}
  S = [(v^S_*, e^S_*), (v^S_{l_t}, e^S_{l_t}), \cdots, (v^S_1, e^S_1)].
\end{equation}

\item \textbf{Operator $\diamond$ application ($\diamond \in \{ +,-,\times,\div \}$)}:
One stack action pops two elements from the top of the stack, which contains two pairs, $(v_i, e_i)$ and $(v_j, e_j)$, and then the associated symbolic operator, $v_k = v_i \diamond v_j$, is recorded.
Also, a semantic transformation function $f_{\diamond}$ for that operator is invoked, which generates the semantic representation of $v_k$ by transforming semantic representations of $v_i$ and $v_j$ to $e_k = f_{\diamond}(e_i, e_j)$.
Therefore, after an operator is applied to the stack specified in (\ref{eq:stack}), the stack state becomes
\begin{align}
\label{eq:apply-op}
  S = & [(v^S_{l_t} \diamond v^S_{l_t-1}, f_{\diamond}(e^S_{l_t}, e^S_{l_t-1})), \\ & (v^S_{l_t-2}, e^S_{l_t-2}), \cdots, (v^S_1, e^S_1)].\nonumber
\end{align}

\item \textbf{Equal application}: When the equal application is chosen, it implies that an equation is completed. 
This stack action pops 2 tuples from the stack, $(v_i, e_i), (v_j, e_j)$, and then $v_i = v_j$ is recorded.
If one of them is an unknown variable, the problem is solved. 
Therefore, after an OP is applied to the stack specified in (\ref{eq:stack}), the stack state becomes
\begin{equation}
  S = [(v^S_{l_t-2}, e^S_{l_t-2}), \cdots, (v^S_1, e^S_1)].
\end{equation}
\end{itemize}

\subsubsection{Operand Selector}
\label{sec:operand-selector}

When the stack action selector has decided to push an operand, the operand selector aims at choosing which operand to push. 
The operand candidates $e$ include constants provided in the problem text whose semantic representations are $e^c_1, e^c_2, \cdots, e^c_n$, unknown variable whose semantic representation is $e^x$, and two external constants $1$ and $\pi$ whose semantic representations are $e^1, e^{\pi}$:
\begin{equation}
  e = [e^c_1, e^c_2, \cdots, e^c_n, e^1, e^\pi, e^x].
\end{equation}
An operand has both symbolic and semantic representations, but the selection focuses on its semantic meaning; this procedure is the same as what human does when solving math word problems.

Inspired by addressing mechanisms of neural Turing machine (NTM)~\cite{graves2014neural}, the probability of choosing the $i$-th operand candidate is the attention weights of $r_t$ over the semantic representations of the operand candidates as in (\ref{eq:attention-weight}):
\begin{align}
    & P(Z_t \mid \{y_i\}_{i=1}^{t-1}, \{ w_i \}_{i=1}^m) \\
  = &\ \mathrm{OperandSelector}(r^{opd}_t) \nonumber\\
  = &\ \mathrm{AttentionWeight}(r^{opd}_t, \{e_i \}_{i=1}^m \cup \{e^1, e^\pi, e^x \} ),\nonumber 
\end{align}
and $r^{opd}_t$ is defined in section \ref{sec:decoding-state-features}.

\subsubsection{Semantic Transformer}
A semantic transformer is proposed to generate the semantic representation of a new symbol resulted from applying an operator, which provides the capability of interpretation and reasoning for the target task.
The semantic transformer for an operator $\diamond \in \{+, -, \times, \div\}$ transforms semantic representations of two operands $e_1, e_2$ into
\begin{equation}
  f_{\diamond}(e_1, e_2) = \tanh(U_{\diamond} \mathrm{ReLU}(W_{\diamond}[e_1; e_2] + b_{\diamond}) + c_{\diamond}),
  \label{eq:semantic-transformer}
\end{equation}
where $W_{\diamond}, U_{\diamond}, b_{\diamond}, c_{\diamond}$ are model parameters.
Semantic transformers for different operators have different parameters in order to model different transformations.

\subsection{Training}
Both stack action selection and operand selection can be trained in a fully supervised way by giving problems and associated ground truth equations.
Because our model generates the equation with stack actions, the equation is first transformed into its postfix representation.
Let the postfix representation of the target equation be $y_1, \cdots y_t, \cdots, y_T$, where $y_t$ can be either an operator ($+, -, \times, \div, =$) or a target operand. 
Then for each time step $t$, the loss can be computed as
\begin{align}L(y_t)&=
  \begin{cases}
    L_1(\mathrm{push\_op}) + L_2(y_t) & \text{$y_t$ is an operand}\\
    L_1(y_t) & \text{otherwise}
  \end{cases},\nonumber
\end{align}
where $L_1$ is the stack action selection loss and $L_2$ is the operand selection loss defined as
\begin{align}
  L_1(y_t) &= - \log P(Y_t = y_t \mid \{o_i\}_{i=1}^{t-1}, \{w_i \}_{i=1}^m),\nonumber\\
  L_2(y_t) &= - \log P(Z_t = y_t \mid r_t).\nonumber
\end{align}
The objective of our training process is to minimize the total loss for the whole equation, $\sum_{t=1}^T L(y_t)$.

\subsection{Inference}

When performing inference, at each time step $t$, the stack action with the highest probability $P(Y_t| \{ \tilde{y}_i \}_{i=1}^{t-1}, \{ w_i \}_{i=1}^m)$ is chosen. 
If the chosen stack action is ``\textit{push}'', the operand with the highest probability $P(Z_t | \{ \tilde{Y}_i \}_{i=1}^{t-1}, \{ w_i \}_{i=1}^m )$ is chosen.
When the stack has less than 2 elements, the probability of applying operator $+, -, \times, \div, =$ would be masked out to prevent illegal stack actions, so all generated equations must be legal math expressions.
The decoder decodes until the unknown variable can be solved.
After the equations are generated, a Python package SymPy \cite{sympy} is used to solve the unknown variable.
The inference procedure example is illustrated in Figure~\ref{fig:infer}.
The detailed algorithm can be found in Algorithm~\ref{algo}.

\begin{algorithm}[h]
\begin{algorithmic}
\small
  \Function{SolveProblem}{problem\_text}
      \State $v \gets$ ExtractConstants(problem\_text)
      \State \ \Comment {$v$ is a list of constants in the problem.}
      \State $h^E, h^D_0, c^D_0, E \gets$ Encoder(problem\_text)
      \State $S \gets \mathrm{Stack}()$
      \State $\mathrm{ret, loss, t, equations} \gets \mathrm{padding}, 0, 1, \{ \}$
      \While {not solvable($\mathrm{equations}$)}
          \State $h^D_t \gets \mathrm{LSTM}(h^D_{t-1}, c_{t-1}, \mathrm{ret})$
          \State $s_t \gets S.\mathrm{get\_top2}()$
          \State $h^E \gets \mathrm{Attention}(h^D_{t-1}, h^E)$
          \State $r_t \gets [h^D_t, s_t, h^E]$
          \State $p_{sa} \gets \mathrm{StackActionSelector}(r_t)$
          \State $p_{opd} \gets \mathrm{OperandSelector}(r_t)$
          \If {training}
              \State \Comment{Target equation $y$ is available when training.}
              \State $Y_t \gets y_t$
              \If {$y_t$ is operand}
                  \State $\mathrm{loss} \gets \mathrm{loss} + L_1(\mathrm{push}) + L_2(y_t)$
              \Else
                  \State $\mathrm{loss} \gets \mathrm{loss} + L_1(y_t)$
              \EndIf
          \Else
              \State $Y_t \gets \mathrm{StackActionSelector}(r_t^{sa})$
              \If {$Y_t = \mathrm{push}$}
                  \State $Z_t \gets \mathrm{OperandSelector}(r_t^{opd})$
              \EndIf
          \EndIf
          \If {$Y_t = \mathrm{gen\_var}$}
              \State $e^x \gets \mathrm{Attention}(h^D_t, h^E)$
              \State $\mathrm{ret} \gets e^x$
          \ElsIf {$Y_t = \mathrm{push}$}
              \State $S.\mathrm{push}(v_{Z_t}, e_{Z_t})$
              \State $\mathrm{ret} \gets e_{Z_t}$
          \ElsIf {$Y_t \in \{ + , - , \times, \div \}$}
              \State $(v_a, e_a), (v_b, e_b) = S.\mathrm{pop}(), S.\mathrm{pop}()$
              \State $S.\mathrm{push}(v_a Y_t v_b, f_{Y_t}(e_{a}, e_{b}) )$
              \State $\mathrm{ret} \gets f_{Y_t}(e_{a}, e_{b})$
          \ElsIf {$Y_t = \mathrm{equal}$}
              \State $(v_a, e_a), (v_b, e_b) = S.\mathrm{pop}(), S.\mathrm{pop}()$
              \State $\mathrm{equations} = \mathrm{equations} \cup "v_a = v_b"$
              \State $\mathrm{ret} \gets S.\mathrm{top}()$
          \EndIf    
      \EndWhile
      \State \Return solve($\mathrm{equations}$)    
  \EndFunction
\end{algorithmic}
\caption{Training and Inference}
\label{algo}
\end{algorithm}

\section{Experiments}
To evaluate the performance of the proposed model, we conduct the experiments on the benchmark dataset and analyze the learned semantics.

\subsection{Settings}
The experiments are benchmarked on the dataset Math23k~ \cite{wang2017deep}, which contains 23,162 math problems with annotated equations. 
Each problem can be solved by a single-unknown-variable equation and only uses operators $+, -, \times, \div$.
Also, except $\pi$ and $1$, quantities in the equation can be found in the problem text. 
There are also other large scale datasets like Dolphin18K \cite{shi2015automatically} and AQuA~\cite{ling2017program}, containing 18,460 and 100,000 math word problems respectively.
The reasons about not evaluating on these two datasets are 1) Dolphin18k contains some unlabeled math word problems and some incorrect labels, and 2) AQuA contains rational for solving the problems, but the equations in the rational are not formal (e.g. mixed with texts, using $x$ to represent $\times$, etc.) and inconsistent.
Therefore, the following experiments are performed and analyzed using Math23K, the only large scaled, good-quality dataset.

\subsection{Results}


The results are shown in Table~\ref{table:accuracy}. 
The retrieval-based methods compare problems in test data with problems in training data, and choose the most similar one's template to solve the problem~\cite{kushman2014learning,upadhyay2017annotating}.
The classification-based models choose equation templates by a classifier trained on the training data. Their performance are reported in \citeauthor{robaidek2018data}. 
The seq2seq and hybrid models are from \citeauthor{wang2017deep}, where the former directly maps natural language into symbols in equations, and the latter one ensembles prediction from a seq2seq model and a retrieval-based model.
The ensemble is the previous state-of-the-art results of Math23K.

\begin{table}[t!]
\begin{tabular*}{\linewidth}{ll@{\extracolsep{\fill}}c}
  \hline
  \multicolumn{2}{l}{\bf Model} & \bf Accuracy \\
  \hline\hline
  \multirow{2}{*}{Retrieval} & Jaccard & 47.2\%    \\
  & Cosine                 & 23.8\%    \\
  \hline
  \multirow{2}{*}{Classification} & BLSTM & 57.9\%      \\
  & Self-Attention & 56.8\%      \\
  \hline
  \multirow{3}{*}{Generation} &  Seq2Seq w/ SNI   & 58.1\%    \\
  & Proposed Word-Based & \textbf{65.3\%}   \\
  & Proposed Char-Based & \textbf{65.8\%}    \\
  \hline
  Hybrid & Retrieval + Seq2Seq & 64.7\%\\
  \hline
\end{tabular*}
\caption{5-fold cross validation results on Math23K.}
\label{table:accuracy}
\end{table}

Our proposed end-to-end model belongs to the generation category, and the single model performance achieved by our proposed model is new state-of-the-art ($>$ 65\%) and even better than the hybrid model result (64.7\%). 
In addition, we are the first to report character-based performance on this dataset, and the character-based results are slightly better than the word-based ones.
Among the single model performance, our models obtain about more than 7\% accuracy improvement compared to the previous best one~\cite{wang2017deep}. The performance of our character-based model also shows that our model is capable of learning the relatively accurate semantic representations without word boundaries and achieves better performance.

\subsection{Ablation Test}

To better understand the performance contributed by each proposed component, we perform a series of ablation tests by removing components one by one and then checking the performance by 5-fold cross validation. Table~\ref{tab:ablation} shows the ablation results.

\paragraph{Char-Based v.s. Word-Based}
As reported above, using word-based model instead of character-based model only causes 0.5\% performance drop.
To fairly compare with prior word-based models, the following ablation tests are performed on the word-based approach.

\begin{table}[t!]
\begin{tabular*}{\linewidth}{l@{\extracolsep{\fill}}c}
  \hline
  \bf Model                         & \bf Accuracy  \\
  \hline
  \hline
  Char-Based                    & 65.8\%    \\
  Word-Based                    & 65.3\%    \\
  Word-Based - Gate             & 64.1\%    \\
  Word-Based - Gate - Attention & 62.5\%    \\
  Word-Based - Gate - Attention - Stack & 60.1\%    \\
  Word-Based - Semantic Transformer & 64.1\% \\
  Word-Based - Semantic Representation & 61.7\% \\
  \hline
\end{tabular*}
\caption{5-fold cross validation results of ablation tests.}
\label{tab:ablation}
\end{table}

\paragraph{Word-Based - Gate}
It uses $r_t$ instead of $r_t^{sa}$ and $r_t^{opr}$ as the input of both $\mathrm{StackActionSelector}$ and $\mathrm{OperandSelector}$. 

\paragraph{Word-Based - Gate - Attention}
Considering that the prior generation-based model (seq2seq) did not use any attention mechanism, we compare the models with and without the attention mechanism.
Removing attention means excluding $q_{t-1}$ in (\ref{eq: operator-gate}), so the input of both operator and operand selector becomes $r_t = [h^D_t; s_t]$.
The result implies that our model is not better than previous models solely because of the attention.

\paragraph{Word-Based - Gate - Attention - Stack} To check the effectiveness of the stack status ($s_{t}$ in (\ref{eq: operator-gate})), the  experiments of removing the stack status from the input of both operator and operand selectors ($r_t = h^D_t$) are conducted.
The results well justify our idea of choosing operators based on semantic meanings of operands.

\paragraph{Word-Based - Semantic Transformer} To validate the effectiveness of the idea that views an operator as a semantic transformer, we modify the semantic transformer function of the operator $\diamond$ into $f_{\diamond}(e_1, e_2) = e_{\diamond}$, where $e_\diamond$ is a learnable parameter and is different for different operators. 
Therefore, $e_\diamond$ acts like the embedding of the operator $\diamond$, and the decoding process is more similar to a general seq2seq model.
The results show that the semantic transformer in the original model encodes not only the last operator applied on the operands but other information that helps the selectors.

\begin{figure*}[t!]
  \centering
  \includegraphics[width=\linewidth]{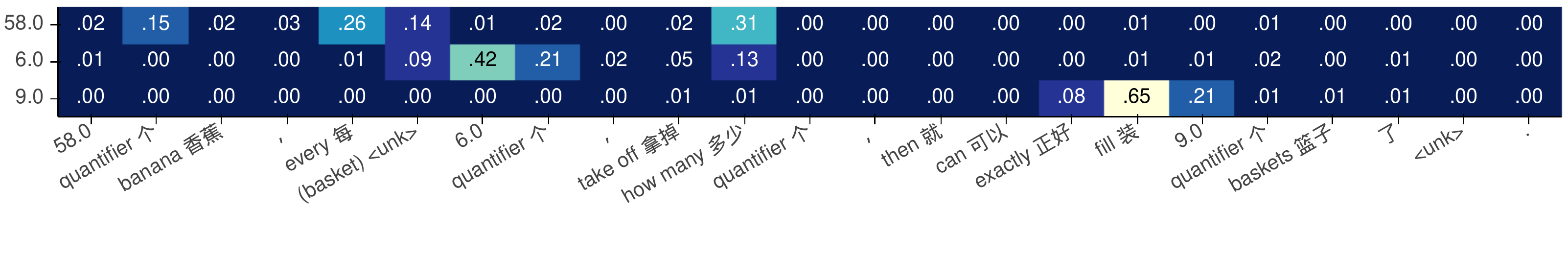}
  \vspace{-9mm}
  \caption{The self-attention map visualization of operands' semantic expressions for the problem ``\textit{There are 58 bananas. Each basket can contain 6 bananas. How many bananas are needed to be token off such that exactly 9 baskets are filled?}''. 
}
  \label{fig:self-attention}
\end{figure*}

\paragraph{Word-Based - Semantic Representation} 
To explicitly evaluate the effectiveness of operands' semantic representations, we 
rewrite semantic representation of the $i$-th operand in the problem texts from (\ref{eq:semantic-representation}) to $e^c_i = b^c_i$, where $b^c_i$ is a parameter.
Thus for every problem, the representation of the $i$-th operand is identical, even though their meanings in different problems may be different.
This modification assumes that no semantic information is captured by $b^c_i$, which can merely represent a symbolic placeholder in an equation.
Because the semantic transformer is to transform the semantic representations, applying this component is meaningless.
Here the semantic transformer is also replaced with $f_{\diamond}(e_1, e_2) = e_{\diamond}$ as the setting of the previous ablation test.
The results show that the model without using semantic representations of operands causes a significant accuracy drop of 3.5\%.
The main contribution of this paper about modeling semantic meanings of symbols is validated and well demonstrated here.

\section{Qualitative Analysis}

To further analyze whether the proposed model can provide interpretation and reasoning, we visualize the learned semantic representations of constants to check where the important cues are, 

\subsection{Constant Embedding Analysis}
To better understand the information encoded in the semantic representations of constants in the problem, a self-attention is performed when their semantic representations are extracted by the encoder. Namely, we rewrite (\ref{eq:semantic-representation}) as
\begin{align}
  e^c_i = \mathrm{Attention}(h_{p_i}^E, \{ h^E_t \}_{t=1}^m.
\end{align}
Then we check the trained self-attention map ($\alpha$ in the attention function) on the validation dataset.

For some problems, the self-attention that generates semantic representations of constants in the problem concentrates on the number's quantifier or unit, and sometimes it also focuses on informative verbs, such as ``\textit{gain}'', ``\textit{get}'', ``\textit{fill}'', etc., in the sentence.
For example, Figure~\ref{fig:self-attention} shows the attention weights for an example math word problem,
where lighter colors indicate higher weights.
The numbers ``\emph{58}'' and ``\emph{6}'' focus more on the quantifier-related words (e.g. ``\emph{every}'' and ``\emph{how many}''), while ``\emph{9}'' pays higher attention to the verb ``\emph{fill}''.
The results are consistent with those hand-craft features for solving math word problems proposed by the prior research~\cite{hosseini2014learning,roy2015solving,roy2015reasoning}.
Hence, we demonstrate that the automatically learned semantic representations indeed  capture critical information that facilitates solving math word problems without providing human-crafted knowledge.

\begin{figure}[t!]
  \centering
  \includegraphics[width=0.95\linewidth]{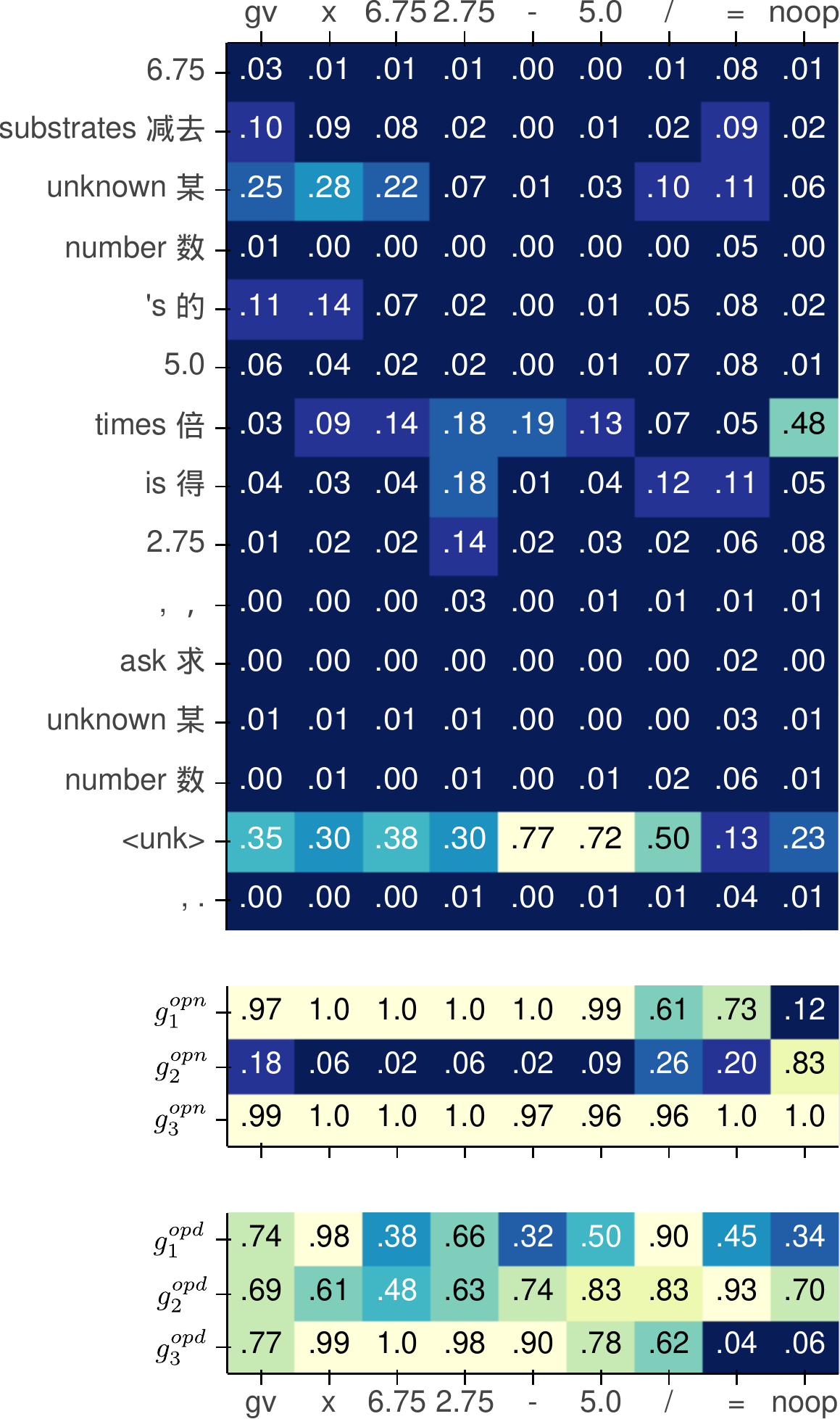}
  \caption{Word attention and gate activation ($g^{sa}$ and $g^{opd}$) visualization when generating stack actions for the problem ``\emph{6.75 deducting 5 times of an unknown number is 2.75. What is the unknown number?}'', where the associated equation is $x=(6.75 - 2.75)\div 5$. Note that $g^{opd}$ is meaningful only when the $t$-th stack action is $\mathrm{push\_op}$.}
  \vspace{-2mm}
  \label{fig:attention}
\end{figure}

\subsection{Decoding Process Visualization}
We visualize the attention map ($q_t$ in (\ref{eq:decoding-attention})) to see how the attention helps the decoding process.
An example is shown in the top of Figure \ref{fig:attention}, where most attention focuses on the end of the sentence.
Unlike the machine translation task, the attention shows the word-level alignment between source and target languages, solving math word problems requires high-level understanding due to the task complexity.

\begin{table*}[t!]
\begin{tabular*}{\linewidth}{p{40em}}
\hline
\bf Problem \& Results\\
\hline\hline
\begin{CJK}{UTF8}{gkai}红花有60朵，黄花比红花多1/6朵，黄花有多少朵．\end{CJK}
(There are 60 red flowers. Yellow flowers are more than red ones by 1/6.
How many yellow flowers are there?)\\
\emph{Generated Equation}: $60 + \frac{1}{6}$\\
\emph{Correct Answer}: 70\\
\hline
\begin{CJK}{UTF8}{gkai}火车 48 小时行驶 5920 千米，汽车 25 小时行驶 2250 千米，汽车平均每小时比火车每小时慢 多少 千米 ？\end{CJK}
(The train travels 5920 kilometers in hours, and the car travels 2250 kilometers in 25 hours. How many kilometers per hour is the car slower than the train?) \\
\emph{Generated Equation}: $2250 \div 25 - 5920 \div 48$\\
\emph{Correct Answer}: $33\frac{1}{3}$\\
\hline
\begin{CJK}{UTF8}{gkai}小红前面 5 人，后面 7 人，一共有多少人？\end{CJK}
(There are 5 people in front of Little Red and 7 people behind.
How many persons are there in total?)\\
\emph{Generated Equation}: $5+7$\\
\emph{Correct Answer}: $13$\\
\hline
\end{tabular*}
\caption{Randomly sampled incorrect predictions.}
\label{tab:error-analysis}
\end{table*}

To further analyze the effectiveness of the proposed gating mechanisms for stack action and operand selection, the activation of gates $g^{sa}, g^{opd}$ at each step of the decoding process is shown in the bottom of Figure~\ref{fig:attention}.
It shows that most of time, the gate activation is high, demonstrating that the proposed gating mechanisms play an important role during decoding.
We also observe a common phenomenon that the activation $g_2^{sa}$, which controls how much attention the stack action selector puts on the stack state when deciding an stack action, is usually low until the last ``operator application'' stack action. 
For example, in the example of Figure~\ref{fig:attention}, $g_2^{sa}$ is less than $0.20$ till the last argument selection stack action, and activates when deciding the \textit{division operator application} ($\div$) and the \textit{equal application} ($=$). 
It may result from the higher-level semantics of the operand $(6.75 - 2.75)$ on the stack when selecting the stack action \textit{division operator application} ($\div$).
In terms of the activation of $g^{opd}$, we find that three features are important in most cases, demonstrating the effectiveness of the proposed mechanisms.

\subsection{Error Analysis}

We randomly sample some results predicted incorrectly by our model shown in Table \ref{tab:error-analysis}.
In the first example, the error is due to the language ambiguity, and such ambiguity cannot be resolved without considering the exact value of the number.
From the second example, although our model identifies the problem as a comparison problem successfully, it handles the order of the operands incorrectly. 
For the third problem, it cannot be solved by using only the surface meaning but requires some common sense. 
Therefore, above phenomena show the difficulty of solving math word problems and the large room for improvement.

\section{Conclusion}

We propose an end-to-end neural math solver using an encoder-decoder framework that incorporates semantic representations of numbers in order to generate mathematical symbols for solving math word problems.
The experiments show that the proposed model achieves the state-of-the-art performance on the benchmark dataset, and empirically demonstrate the effectiveness of each component in the model.
In sum, the proposed neural math solver is designed based on how human performs reasoning when writing equations, providing better interpretation without the need of labeled rationals.

\bibliographystyle{acl_natbib}
\bibliography{naaclhlt2019.bib}

\begin{thebibliography}{23}
\expandafter\ifx\csname natexlab\endcsname\relax\def\natexlab#1{#1}\fi

\bibitem[{Graves et~al.(2014)Graves, Wayne, and Danihelka}]{graves2014neural}
Alex Graves, Greg Wayne, and Ivo Danihelka. 2014.
\newblock Neural turing machines.
\newblock \emph{arXiv preprint arXiv:1410.5401}.

\bibitem[{Hochreiter and Schmidhuber(1997)}]{hochreiter1997long}
Sepp Hochreiter and J{\"{u}}rgen Schmidhuber. 1997.
\newblock Long short-term memory.
\newblock \emph{Neural Computation}, 9(8):1735--1780.

\bibitem[{Hosseini et~al.(2014)Hosseini, Hajishirzi, Etzioni, and
  Kushman}]{hosseini2014learning}
Mohammad~Javad Hosseini, Hannaneh Hajishirzi, Oren Etzioni, and Nate Kushman.
  2014.
\newblock Learning to solve arithmetic word problems with verb categorization.
\newblock In \emph{Proceedings of the 2014 Conference on Empirical Methods in
  Natural Language Processing}, pages 523--533.

\bibitem[{Joulin et~al.(2016)Joulin, Grave, Bojanowski, Douze, J{\'e}gou, and
  Mikolov}]{joulin2016fasttext}
Armand Joulin, Edouard Grave, Piotr Bojanowski, Matthijs Douze, H{\'e}rve
  J{\'e}gou, and Tomas Mikolov. 2016.
\newblock Fasttext.zip: Compressing text classification models.
\newblock \emph{arXiv preprint arXiv:1612.03651}.

\bibitem[{Kingma and Ba(2014)}]{kingma2014adam}
Diederik~P. Kingma and Jimmy Ba. 2014.
\newblock Adam: {A} method for stochastic optimization.
\newblock \emph{CoRR}, abs/1412.6980.

\bibitem[{Koncel{-}Kedziorski et~al.(2015)Koncel{-}Kedziorski, Hajishirzi,
  Sabharwal, Etzioni, and Ang}]{koncel2015parsing}
Rik Koncel{-}Kedziorski, Hannaneh Hajishirzi, Ashish Sabharwal, Oren Etzioni,
  and Siena~Dumas Ang. 2015.
\newblock Parsing algebraic word problems into equations.
\newblock \emph{{TACL}}, 3:585--597.

\bibitem[{Kushman et~al.(2014)Kushman, Zettlemoyer, Barzilay, and
  Artzi}]{kushman2014learning}
Nate Kushman, Luke Zettlemoyer, Regina Barzilay, and Yoav Artzi. 2014.
\newblock Learning to automatically solve algebra word problems.
\newblock In \emph{Proceedings of the 52nd Annual Meeting of the Association
  for Computational Linguistics, {ACL} 2014}, pages 271--281.

\bibitem[{Ling et~al.(2017)Ling, Yogatama, Dyer, and Blunsom}]{ling2017program}
Wang Ling, Dani Yogatama, Chris Dyer, and Phil Blunsom. 2017.
\newblock Program induction by rationale generation: Learning to solve and
  explain algebraic word problems.
\newblock In \emph{Proceedings of the 55th Annual Meeting of the Association
  for Computational Linguistics, {ACL} 2017}, pages 158--167.

\bibitem[{Luong et~al.(2015)Luong, Pham, and Manning}]{luong2015effective}
Thang Luong, Hieu Pham, and Christopher~D. Manning. 2015.
\newblock Effective approaches to attention-based neural machine translation.
\newblock In \emph{Proceedings of the 2015 Conference on Empirical Methods in
  Natural Language Processing}, pages 1412--1421.

\bibitem[{Mandal and Naskar(2019)}]{mandal2019solving}
Sourav Mandal and Sudip~Kumar Naskar. 2019.
\newblock Solving arithmetic mathematical word problems: A review and recent
  advancements.
\newblock In \emph{Information Technology and Applied Mathematics}, pages
  95--114. Springer.

\bibitem[{Mehta et~al.(2017)Mehta, Mishra, Athavale, Shrivastava, and
  Sharma}]{mehta2017deep}
Purvanshi Mehta, Pruthwik Mishra, Vinayak Athavale, Manish Shrivastava, and
  Dipti~Misra Sharma. 2017.
\newblock Deep neural network based system for solving arithmetic word
  problems.
\newblock In \emph{Proceedings of the {IJCNLP} 2017}, pages 65--68.

\bibitem[{Meurer et~al.(2017)Meurer, Smith, Paprocki, \v{C}ert\'{i}k,
  Kirpichev, Rocklin, Kumar, Ivanov, Moore, Singh, Rathnayake, Vig, Granger,
  Muller, Bonazzi, Gupta, Vats, Johansson, Pedregosa, Curry, Terrel,
  Rou\v{c}ka, Saboo, Fernando, Kulal, Cimrman, and Scopatz}]{sympy}
Aaron Meurer, Christopher~P. Smith, Mateusz Paprocki, Ond\v{r}ej
  \v{C}ert\'{i}k, Sergey~B. Kirpichev, Matthew Rocklin, AMiT Kumar, Sergiu
  Ivanov, Jason~K. Moore, Sartaj Singh, Thilina Rathnayake, Sean Vig, Brian~E.
  Granger, Richard~P. Muller, Francesco Bonazzi, Harsh Gupta, Shivam Vats,
  Fredrik Johansson, Fabian Pedregosa, Matthew~J. Curry, Andy~R. Terrel,
  \v{S}t\v{e}p\'{a}n Rou\v{c}ka, Ashutosh Saboo, Isuru Fernando, Sumith Kulal,
  Robert Cimrman, and Anthony Scopatz. 2017.
\newblock \href {https://doi.org/10.7717/peerj-cs.103} {Sympy: symbolic
  computing in python}.
\newblock \emph{PeerJ Computer Science}, 3:e103.

\bibitem[{Robaidek et~al.(2018)Robaidek, Koncel{-}Kedziorski, and
  Hajishirzi}]{robaidek2018data}
Benjamin Robaidek, Rik Koncel{-}Kedziorski, and Hannaneh Hajishirzi. 2018.
\newblock Data-driven methods for solving algebra word problems.
\newblock \emph{CoRR}, abs/1804.10718.

\bibitem[{Roy and Roth(2015)}]{roy2015solving}
Subhro Roy and Dan Roth. 2015.
\newblock Solving general arithmetic word problems.
\newblock In \emph{Proceedings of the 2015 Conference on Empirical Methods in
  Natural Language Processing, {EMNLP} 2015, Lisbon, Portugal, September 17-21,
  2015}, pages 1743--1752.

\bibitem[{Roy and Roth(2018)}]{roy2018mapping}
Subhro Roy and Dan Roth. 2018.
\newblock Mapping to declarative knowledge for word problem solving.
\newblock \emph{{TACL}}, 6:159--172.

\bibitem[{Roy et~al.(2016)Roy, Upadhyay, and Roth}]{roy2016equation}
Subhro Roy, Shyam Upadhyay, and Dan Roth. 2016.
\newblock Equation parsing : Mapping sentences to grounded equations.
\newblock In \emph{Proceedings of the 2016 Conference on Empirical Methods in
  Natural Language Processing, {EMNLP} 2016, Austin, Texas, USA, November 1-4,
  2016}, pages 1088--1097.

\bibitem[{Roy et~al.(2015)Roy, Vieira, and Roth}]{roy2015reasoning}
Subhro Roy, Tim Vieira, and Dan Roth. 2015.
\newblock Reasoning about quantities in natural language.
\newblock \emph{{TACL}}, 3:1--13.

\bibitem[{Shi et~al.(2015)Shi, Wang, Lin, Liu, and Rui}]{shi2015automatically}
Shuming Shi, Yuehui Wang, Chin{-}Yew Lin, Xiaojiang Liu, and Yong Rui. 2015.
\newblock Automatically solving number word problems by semantic parsing and
  reasoning.
\newblock In \emph{Proceedings of the 2015 Conference on Empirical Methods in
  Natural Language Processing, {EMNLP} 2015}, pages 1132--1142.

\bibitem[{Sutskever et~al.(2014)Sutskever, Vinyals, and
  Le}]{sutskever2014sequence}
Ilya Sutskever, Oriol Vinyals, and Quoc~V. Le. 2014.
\newblock Sequence to sequence learning with neural networks.
\newblock In \emph{Advances in Neural Information Processing Systems 27: Annual
  Conference on Neural Information Processing Systems 2014}, pages 3104--3112.

\bibitem[{Upadhyay and Chang(2017)}]{upadhyay2017annotating}
Shyam Upadhyay and Ming{-}Wei Chang. 2017.
\newblock Annotating derivations: {A} new evaluation strategy and dataset for
  algebra word problems.
\newblock In \emph{Proceedings of the 15th Conference of the European Chapter
  of the Association for Computational Linguistics}, pages 494--504.

\bibitem[{Upadhyay et~al.(2016)Upadhyay, Chang, Chang, and
  Yih}]{upadhyay2016learning}
Shyam Upadhyay, Ming{-}Wei Chang, Kai{-}Wei Chang, and Wen{-}tau Yih. 2016.
\newblock Learning from explicit and implicit supervision jointly for algebra
  word problems.
\newblock In \emph{Proceedings of the 2016 Conference on Empirical Methods in
  Natural Language Processing}, pages 297--306.

\bibitem[{Wang et~al.(2018)Wang, Zhang, Gao, Song, Guo, and
  Shen}]{wang2018mathdqn}
Lei Wang, Dongxiang Zhang, Lianli Gao, Jingkuan Song, Long Guo, and Heng~Tao
  Shen. 2018.
\newblock {MathDQN}: Solving arithmetic word problems via deep reinforcement
  learning.
\newblock In \emph{Proceedings of the Thirty-Second {AAAI} Conference on
  Artificial Intelligence}.

\bibitem[{Wang et~al.(2017)Wang, Liu, and Shi}]{wang2017deep}
Yan Wang, Xiaojiang Liu, and Shuming Shi. 2017.
\newblock Deep neural solver for math word problems.
\newblock In \emph{Proceedings of the 2017 Conference on Empirical Methods in
  Natural Language Processing}, pages 845--854.

\end{thebibliography}

\appendix

\section{Algorithm Detail}
The training and inference procedures are shown in Algortihm~\ref{algo}.

\section{Hyperparameter Setup}

The model is trained with the optimizer \texttt{adam} \cite{kingma2014adam}, and the learning rate is set to $0.001$. Pretrained embeddings using FastText \cite{joulin2016fasttext} are adopted.
The hidden state size of LSTM used in the encoder and decoder is 256. The dimension of hidden layers in attention, semantic transformer and operand/stack action selector is 256.
The dropout rate is set as $0.1$  before inputting the decoder LSTM, before the stack action selector and after the hidden layer of the stack action selector and attention. 
The reported accuracy is the result of 5-fold cross-validation, same as \citeauthor{wang2017deep} for fair comparison.

\begin{table*}[!htbp]
\begin{tabular*}{\linewidth}{p{40em}}
\hline
\bf Problem \& Results\\
\hline\hline
\begin{CJK}{UTF8}{gkai}小红前面 5 人，后面 7 人，一共有多少人？\end{CJK}
(There are 5 people in front of Little Red and 7 people behind.
  How many persons are there in total?)\\
\emph{Proposed Model}: $5+7$ \\
\emph{Seq2Seq Model}: $5+7+1$ \\
\hline
\begin{CJK}{UTF8}{gkai}两个数相差28，如果被减数减少3，减数增加5，那么它们的差=？\end{CJK}
(The difference between two numbers is 28. If the minuend is reduced by 3, and
  the subtrahend is increased by 5, then their difference=?)\\
\emph{Proposed Model}: $(28-3) \div 5$ \\
\emph{Seq2Seq Model}: $28-(3+5)$ \\
\hline
\begin{CJK}{UTF8}{gkai}机床厂第一车间有55人，第二车间有45人，每人每天平均生产261个零件，这两个
  车间每天共生产多少个零件？\end{CJK}
(There are 55 people in the first workshop of the machine tool
  factory and 45 people in the second workshop. Each person
  produces 261 small components per day in average. How many components do the
  two workshops produce every day in total?)\\
\emph{Proposed Model}: $(55+45) \div 261$ \\
\emph{Seq2Seq Model}: $(55+45) \times 261$ \\
\hline
\begin{CJK}{UTF8}{gkai}箭鱼游动时的速度是28米/秒，8秒可以游多少米？\end{CJK}
(The swordfish swims at speed 28 meters/sec. How many meters
  can it swim in 8 seconds?)\\
\emph{Proposed Model}: $28 \div 8$ \\
\emph{Seq2Seq Model}: $28 \times 8$ \\
\hline
\begin{CJK}{UTF8}{gkai}水果店有梨子387千克，卖出205千克后，又运来一批，现在水果店共有梨子945千克．水果店又运来梨子多少千克？\end{CJK}
(The fruit shop has 387 kilograms of pears . After selling 205 kilograms, some pears arrive. Now the fruit shop has 945 kilograms of pears in total. How many kilograms of pears does the fruit shop get?)\\
\emph{Proposed Model}: $945 \times (387 - 205)$ \\
\emph{Seq2Seq Model}: $945-(387-205)$ \\
\hline
\begin{CJK}{UTF8}{gkai}王老师买排球用了40元，买篮球用的钱数是排球的3倍．王老师买球一共用了多少元？\end{CJK}
(Teacher Wang spent 40 dollars buying volleyballs and 3 times of money for basketballs. How many dollars did Teacher Wang spend for the balls?) \\
\emph{Proposed Model}: $40 \div 3 + 40$ \\
\emph{Seq2Seq Model}: $40 + 40 \times 3$ \\
\hline
\begin{CJK}{UTF8}{gkai}筑路队修筑一条长1200米的公路，甲队单独修40天可以完成任务，乙队单独修30天可以完成任务．甲队每天修的比乙队少多少米？\end{CJK}
(The road construction team built a road with a length of 1200 meters. Team A can complete the task in 40 days alone, and team B can complete the task in 30 days alone. How many meters does team A construct more than team B every day?)\\
\emph{Proposed Model}: $1200 \div 40 - 1200 \div 30$ \\
\emph{Seq2Seq Model}: $1200 \div 30 - 1200 \div 40$ \\
\hline
\begin{CJK}{UTF8}{gkai}一共1800本，我们六年级分得2/9，分给五年级的本数相当于六年级的4/5，五年级分得多少本？\end{CJK}
(There are 1800 books in total. We sixth grade get 2/9. The number of books given to the fifth grade is equal to 4/5 of the number to the sixth grade. How many books does the fifth grade get?)\\
\emph{Proposed Model}: $1800 \times \frac{2}{9} \div \frac{4}{5}$ \\
\emph{Seq2Seq Model}: $1800 \times \frac{2}{9} \times \frac{4}{5}$ \\
\hline
\begin{CJK}{UTF8}{gkai}有一批布料，如果只做上衣可以做10件，如果只做裤子可以做15条，那么这批布料可以做几套这样的衣服？\end{CJK}
(There is a batch of fabrics. If all is used for making shirts, 10 pieces can be made, and 15 pieces if used to make pants only. Then how many suits of such clothes can be made with this batch of fabric?)\\
\emph{Proposed Model}: $10 \times 1 \div 15$ \\
\emph{Seq2Seq Model}: $1 \div (1 \div 10 + 1 \div 15)$ \\
\hline
\begin{CJK}{UTF8}{gkai}贝贝的钱买一本5.9元笔记本差0.6元，他买一本4.8元的，剩下的钱正好买一只圆珠笔，这只圆珠笔多少钱？\end{CJK}
(Beibei needs 0.6 dollars more to buy a notebook of 5.9 dollars. If he buys one of 4.8 dollars, the remaining money allows her to buy exactly one ball pen. How much is the ball pen?)\\
\emph{Proposed Model}: $5.9+0.6-4.8$ \\
\emph{Seq2Seq Model}: $5.9-0.6-4.8$ \\
\hline
\end{tabular*}
\caption{Examples that Seq2Seq predicts correctly while our proposed model predicts incorrectly.}
\label{tab:seq2seq-correct}
\end{table*}

\begin{table*}[t!]
\begin{tabular*}{\linewidth}{p{40em}}
\hline
\bf Problem \& Results\\
\hline\hline
\begin{CJK}{UTF8}{gkai}医院里经常要给病人输入葡萄糖水，这种葡萄糖水是把葡萄糖和水按1：19配制的，根据这些信息，你能知道什么？\end{CJK}
(In hospital, it is often necessary to give glucose injection to patient. This glucose water is prepared by mixing glucose and water at 1:19. Based on this information, what do you know?)\\
\emph{Proposed Model}: $1 \div (1 + 19.0)$ \\
\emph{Seq2Seq Model}: $1 \times (1 + 19.0)$ \\
\hline
\begin{CJK}{UTF8}{gkai}一根长2.45米的木桩打入河底，现在测得木桩水上部分长0.75米，水中长1.05米，求这根桩打在泥中的长度=多少米？\end{CJK}
(A wooden pile of 2.45 meters long is hammered into the bottom of a river. Now the part above water is measured as 0.75 meters long, and the part in the water is measured as 1.05 meters long. How long is the part of the pile in the mud?)\\
\emph{Proposed Model}: $2.45 - 0.75 - 1.05$ \\
\emph{Seq2Seq Model}: $2.45+0.75+1.05$ \\
\hline
\begin{CJK}{UTF8}{gkai}李强6月份的生活费为255元，比计划节省了15\%，节省了多少元．\end{CJK}
(Li Qiang's living expenses in June were 255 dollars, 15\% savings over the plan. How much did he save?)\\
\emph{Proposed Model}: $(255.0 \div (1 - 0.15)) \times 0.15$ \\
\emph{Seq2Seq Model}: $0.15=6.0/(1-255.0)-6.0$ \\
\hline
\begin{CJK}{UTF8}{gkai}小芳在计算一个数除以10时，将除号看成了乘号，结果得3.2，正确的结果应该=．\end{CJK}
(When Xiaofang calculates a number divided by 10 , the division sign is mistakenly treated as a multiplication sign, and the result is 3.2. The correct result should be = .)\\
\emph{Proposed Model}: $3 \div 10 \div 10$ \\
\emph{Seq2Seq Model}: $3.2 \div (1+10)$ \\
\hline
\begin{CJK}{UTF8}{gkai}24 + 91 的 2/13，所得的和再除 19/20，商 = ？\end{CJK}
(2/13 of 91 + 24, and the sum is divided by 19/20, quotient = ?)\\
\emph{Proposed Model}: $\frac{19}{20} \div (24 + 91 \times \frac{2}{13})$ \\
\emph{Seq2Seq Model}: $\frac{19}{20} \div (24 \times 91 - \frac{2}{13})$ \\
\hline

\begin{CJK}{UTF8}{gkai}1/3 + 0.25 = ？\end{CJK}
(1/3 + 0.25 = ?) \\
\emph{Proposed Model}: $\frac{1}{3} + 0.25$ \\
\emph{Seq2Seq Model}: $\frac{1}{3} \times 0.25$ \\
\hline
\begin{CJK}{UTF8}{gkai}商店运来鸡蛋和鸭蛋各7箱．鸡蛋每箱重26千克，鸭蛋每箱重31千克，商店一共运来的鸡蛋和鸭蛋共多少千克？\end{CJK}
(The store shipped 7 boxes of eggs and duck eggs respectively. Eggs weigh 26 kilograms per box, duck eggs weigh 31 kilograms per box. How many kilograms of eggs and duck eggs are shipped from the store in total?)\\
\emph{Proposed Model}: $26 \times 7 + 31 \times 7$ \\
\emph{Seq2Seq Model}: $26 \times 7 + 31$ \\
\hline

\begin{CJK}{UTF8}{gkai}3.8 - 2.54 + 1.46 = ？\end{CJK}
(3.8 - 2.54 + 1.46 =) \\
\emph{Proposed Model}: $3.8 - 2.54 + 1.46$ \\
\emph{Seq2Seq Model}: $3.8+2.54+1.46$ \\
\hline

\begin{CJK}{UTF8}{gkai}有一池水，第一天放出200吨，第二天比第一天多放20\%，第3天放了整池水的36\%，正好全部放完．这池水共
  有多少吨？\end{CJK}
(There was a pool of water, which released 200 tons of water in the first day, 20\% more in the second day than the first day, and 36\% of the whole pool on the third day. Then the water is gone.
  How many tons of water did this pool have?)\\
\emph{Proposed Model}: $(200.0 + 200.0 \times (1 + 0.2)) \div (1 - 0.36)$ \\
\emph{Seq2Seq Model}: $(200.0+0.2) \times 3.0+0.2 \times (1-0.36)$ \\
\hline
\begin{CJK}{UTF8}{gkai}16 的 5/12 比一个数的 7 倍多 2 ， 这个数 = ？\end{CJK}
(5/12 of 16 is more than 7 times of a number by 2. What is the number=?)\\
\emph{Proposed Model}: $(16 \times \frac{5}{12} - 2) \div 7$ \\
\emph{Seq2Seq Model}: $(16 \times \frac{5}{12} + 7) \div 2$ \\
\hline

\end{tabular*}
\caption{Examples that Seq2Seq predicts incorrectly while our proposed model predicts correctly.}
\label{tab:seq2seq-incorrect}
\end{table*}

\section{Error Analysis between Seq2Seq}
We implement the seq2seq model as proposed by \citeauthor{wang2017deep} and compare the performance difference between our proposed model and the baseline seq2seq model.
Table \ref{tab:seq2seq-correct} shows the generated results seq2seq predicts correctly but our model predicts incorrectly.
Table \ref{tab:seq2seq-incorrect} show the results our model can predict correctly but seq2seq cannot.

\end{document}